\documentclass{article} 
\usepackage{iclr2026_conference,times}
\usepackage{mathtools}

\usepackage{amsmath,amsfonts,bm}

\def\eqref#1{equation~\ref{#1}}

\def\1{\bm{1}}

\DeclareMathAlphabet{\mathsfit}{\encodingdefault}{\sfdefault}{m}{sl}
\SetMathAlphabet{\mathsfit}{bold}{\encodingdefault}{\sfdefault}{bx}{n}

\usepackage{amsthm}
\usepackage{booktabs} 
\usepackage{float}
\usepackage{hyperref}
\usepackage{url}
\usepackage{algorithmic}
\usepackage[table]{xcolor} 
\definecolor{navy}{rgb}{0.0,0.0,0} 
\definecolor{lightblue}{RGB}{50,85,210}

\definecolor{lightpink}{RGB}{128,0,32} 

\usepackage{algorithm}
\usepackage{footmisc}

\usepackage{amsthm}

\usepackage[most]{tcolorbox}
\tcbuselibrary{theorems}

\tcolorboxenvironment{proposition}{
  colback=blue!5!white,    
  colframe=blue!75!black,  
  fonttitle=\bfseries,
  coltitle=black,
  sharp corners,
  boxrule=0.8pt,
  left=2mm,
  right=2mm,
  top=1mm,
  bottom=1mm,
  enhanced,
}

\tcbuselibrary{theorems}

\newcommand{\highlight}[1]{\textbf{\textcolor{blue!55!black}{#1}}}
\makeatletter
\renewcommand\@fnsymbol[1]{}  
\makeatother

\title{Latent-DARM: Bridging Discrete Diffusion And Autoregressive Models For Reasoning}

\author{
\begin{minipage}{\textwidth}
\centering
\textbf{Lina Berrayana}$^{1\dagger\ddagger}$, \textbf{Ahmed Heakl}$^{2\dagger}$, \textbf{Muhammad Abdullah Sohail}$^{2}$, \textbf{Thomas Hofmann}$^{3}$, \textbf{Salman Khan}$^{2}$, \textbf{Wei Chen}$^{4\ddagger}$ \\
{\normalfont $^{1}$EPFL \qquad $^{2}$MBZUAI \qquad $^{3}$ETH Zürich \qquad $^{4}$Microsoft Research Asia}
\end{minipage}
\thanks{\begin{tabular}[t]{@{}l@{\,}l}$^\dagger$ & Equal contribution.\\ $^\ddagger$ & Corresponding authors: \texttt{weic@microsoft.com, lina.berrayana@epfl.ch}\end{tabular}}
}

\iclrfinalcopy 
\begin{document}

\maketitle

\begin{abstract}
Most multi-agent systems rely exclusively on autoregressive language models (ARMs) that are based on sequential generation. Although effective for fluent text, ARMs limit global reasoning and plan revision. On the other hand, Discrete Diffusion Language Models (DDLMs) enable non-sequential, globally revisable generation and have shown strong planning capabilities, but their limited text fluency hinders direct collaboration with ARMs. 
We introduce \textbf{Latent-DARM}, a latent-space communication framework bridging DDLM (planners) and ARM (executors), maximizing collaborative benefits. 
Across mathematical, scientific, and commonsense reasoning benchmarks, Latent-DARM outperforms text-based interfaces on average, improving accuracy from 27.0\% to 36.0\% on DART-5 and from 0.0\% to 14.0\% on AIME~2024. Latent-DARM approaches the results of state-of-the-art reasoning models while using less than 2.2\% of its token budget. 
This work advances multi-agent collaboration among agents with heterogeneous models.
\end{abstract}

\section{Introduction}

Collaborative interactions between models are increasingly recognized as a fundamental driver of system-level intelligence in the era of agentic AI \citep{acharya2025agentic, guo2402large}. Advances in multi-agent systems (MAS) research \citep{hong2023metagpt, chen2023autoagents, wu2024autogen} have shifted the prevailing paradigm away from isolated, single-model reasoning toward intelligence that emerges through coordination among multiple agents, often with complementary and specialized roles. Within this setting, MAS built on large language models (LLMs) have demonstrated strong performance in a wide range of applications, such as collaborative reasoning in mathematics and scientific problem solving \citep{karbasi2025multi, zhang2025debate4math}, as well as common-sense reasoning \citep{panzarasa2002formalizing}.

Despite this progress, most existing MAS rely exclusively on autoregressive language models (ARMs), which generate output token by token in a strictly sequential manner. As a consequence, every stage of the agentic workflow remains inherently autoregressive. 
In contrast, discrete diffusion language models (DDLMs) \citep{sahoo2024simple, gat2024discrete, shi2024simplified} have recently attracted increasing attention, driven by evidence that they can outperform ARMs on complex reasoning and planning tasks \citep{ye2024diffusion}. Therefore, relying solely on ARMs may constrain the full potential of MAS, particularly for tasks that require flexible planning or global reasoning.

\begin{figure*}
    \centering
\includegraphics[width=0.6\linewidth]{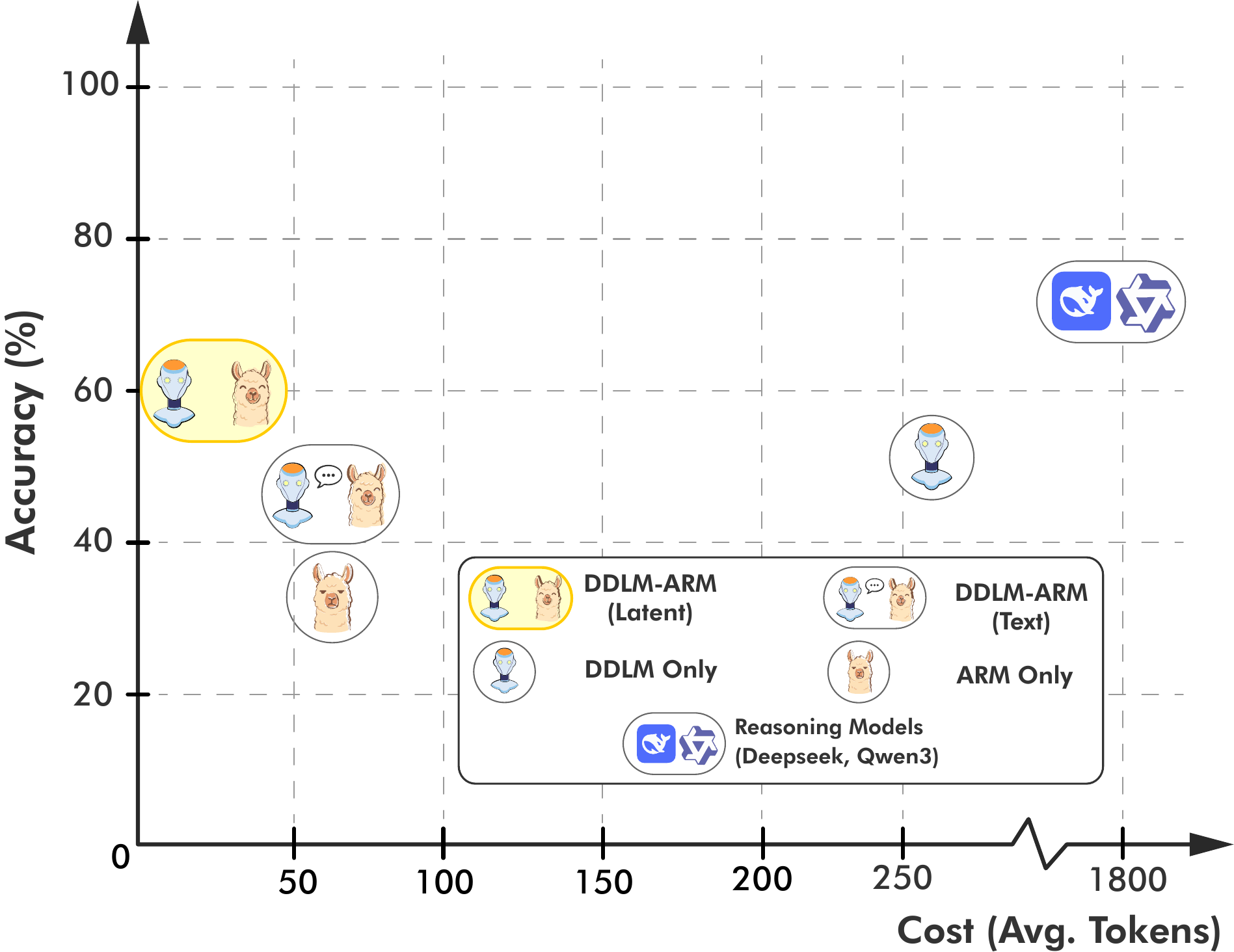}
    \caption{ Accuracy–cost trade-offs across planner–executor configurations. Here DDLM refers to a LLada-8B-Instruct and ARM to a  Llama-3.2-3B-Instruct model.
    DDLM$\to$ARM, particularly with latent-space exchange, achieves higher reasoning accuracy 
    at lower token budgets compared to ARM-only models.}
    \label{fig:teaser}
\end{figure*}

However, a key limitation remains: DDLMs still lag behind ARMs in terms of text fluency. This gap can hinder effective communication between the two agents, particularly when the DDLM-generated output lacks sufficient linguistic coherence. This observation naturally raises the following question: 
\begin{quote}
\emph{\textcolor{blue!55!black}{How to take advantage of DDLMs and ARMs properties while optimizing communication between the two models ?}}
\end{quote}
This work presents a preliminary empirical investigation of this question, as illustrated in Fig. \ref{fig:teaser}. We introduce \highlight{Latent-DARM} (\highlight{Latent}- Discrete \highlight{D}iffusion and \highlight{A}uto\highlight{R}egressive \highlight{M}odel Communication), a communication framework that bridges DDLMs and ARMs operating within the latent space. For that, we study a planner–executor framework in which a DDLM generates a solution plan for a given problem and an autoregressive model (ARM) executes the plan to produce the final answer. We empirically evaluated the effectiveness of this latent-space communication against traditional text-based collaboration across benchmarks in mathematical and scientific reasoning, as well as commonsense understanding.

\paragraph{Why consider a planner–executor framework?}
First, this setting is quite common in multi-agent systems \citep{he2025plan}. Additionally, key advantage of DDLMs is their ability to generate tokens in arbitrary orders, enabling non-sequential and bidirectional planning that more closely aligns with human reasoning processes. In mathematical problem solving, for instance, humans often reason backward from the desired solution, iteratively refining intermediate steps before presenting a fully sequential proof. Similarly, commonsense reasoning proceeds non-linearly, with salient considerations identified prior to explicit verbalization. Once such a plan is formed, it is subsequently articulated in a sequential linguistic form.  
Motivated by this analogy, we adopt a planner--executor multi-agent framework as our experimental setting to investigate the central research question.

\paragraph{Contributions.} 
Our main contributions are twofold:
\begin{itemize}
    \item We provide empirical insights into collaborations between DDLMs and ARMs in a MAS context.
    \item We introduce a latent-based communication framework to improve DDLMs and ARMs integration performance. To our knowledge, we introduce the first latent-space communication solution designed to bridge models with fundamentally different architectures and latent representations.
\end{itemize}

\paragraph{Organization.} 
We first review the relevant definitions and preliminaries and motivate our work in Section 2. In Section 3, we introduce Latent-DARM and describe how it addresses the challenge of bridging DDLMs and ARMs. In Section 4, we present the experimental details. Finally, we discuss the results in Section 5 and show that the observed accuracy improvements stem from a better processing of the plan.

\section{Preliminaries}

\subsection{Definitions}
In our framework, we define two complementary roles: the \emph{planner} and the \emph{executor}.

\textbf{Planner:} Planner is a language model that is responsible for generating intermediate outputs—such as plans, hints, or key facts—that guide the reasoning process without directly producing the final answer. This phase corresponds to the ``thinking'' stage, distinct from the final response generation. Following the \emph{Plan-and-Solve} prompt method~\citep{wang2023planandsolvepromptingimprovingzeroshot}, which improves reasoning performance, the planner formulates the solution plan.

\textbf{Executor:} Executor is a language model that generates the final answer leveraging both the original query and the planner’s output, without engaging in further explicit reasoning.

In the multi-agent system (MAS) context, we refer to the planner as \emph{Agent 1} and the executor as \emph{Agent 2}.

\subsection{Motivation}




In this work, we focus on DDLM, specifically Masked Discrete Diffusion Language Models~\citep{sahoo2024simple} as \emph{Agent 1} and ARMs as \emph{Agent 2}. 

\paragraph{DDLMs for planning.}  
Through their iterative denoising process, DDLMs enable flexible, non-sequential token generation, allowing the model to condition on global context when constructing sequences. This property makes diffusion-based models particularly attractive for planning and structured reasoning tasks~\citep{ye2024beyond}.
In contrast, ARMs, trained to generate sequences via prediction of next-tokens from left-to-right, force decisions to be made based on prefixes generated previously, restricting the model’s ability to review previous choices or reason over the global structure.
As shown by \citet{bachmann2024pitfalls}, ARMs can learn spurious heuristics that exploit prefix-level shortcuts rather than internalizing the underlying planning dynamics, thereby limiting their effectiveness in certain planning tasks.

Recent work on hybrid approaches provides encouraging evidence for the collaboration between these paradigms. \citet{arriola2025block} introduce Block Diffusion models that interpolate between autoregressive and diffusion generation, achieving state-of-the-art performance among diffusion models while supporting flexible-length generation. Their results demonstrate that combining the strengths of both approaches --sequential coherence from ARMs and parallel generation from diffusion---can overcome the limitations of either method alone.

\paragraph{Fluency Problem.}  
However, a key limitation of DDLMs that must be addressed in such collaborative frameworks is output fluency, as disfluent outputs can degrade the quality of the messages sent to the next agent. For example, perplexity, which is often used as a proxy for fluency \cite{feng2025theoreticalbenefitlimitationdiffusion}, is generally higher for DDLMs, indicating worse fluency, compared to state-of-the-art ARMs \cite{sahoo2024simple}. Let $\mathbf{x} \in \{1,\dots,V\}^L$ denote a sequence of length $L$ in vocabulary $V$.  During training, DDLMs learn to model the \textbf{unmasking posterior} $p_\theta(x_i = \cdot \mid z)$ for each masked position $i$ in a partially masked sequence $z$. The masking process samples $t \sim \text{Unif}[0,1]$ and replaces each token $x_i$ with a mask token $m$ independently of probability $t$, creating a corrupted sequence $z$. The model $p_\theta$ is trained to minimize:
\begin{equation}
\mathbb{E}_{\mathbf{x} \sim q_{\text{data}}, t, z} \left[\sum_{i : z_i = m} -\log p_\theta(x_i \mid z) \right],
\end{equation}
where the expectation is over data sequences $\mathbf{x} \sim q_{\text{data}}$, mask ratios $t$, and masking patterns $z$~\citep{sahoo2024simple,shi2024simplified}. 
Generation reverses this corruption: starting from fully masked input, the model iteratively unmasks tokens through multiple denoising steps, until producing clean output. However, tokens revealed simultaneously at a given denoising step are predicted \textbf{independently} given only the current unmasked context, which can undermine fluency \cite{feng2025theoreticalbenefitlimitationdiffusion}.
Recent work has sought to address this issue by deriving optimal unmasking schedules that minimize sampling error through connections to function approximation theory \cite{chen2025optimal}, or by introducing self-correction mechanisms for diffusion-based generation \cite{von2025generalized, kim2025fine}.
Our work takes a \textbf{complementary approach}: rather than solely addressing fluency through improved diffusion techniques, we propose a multi-agent framework where DDLMs generate high-level plans leveraging their global reasoning capabilities, while ARMs execute these plans with sequential fluency. 

\paragraph{Remark.} This division of labor mirrors \textbf{human cognitive processes} --internal flexible thinking followed by sequential articulation---and allows each model to operate within its strengths while mitigating its weaknesses.

\subsection{Problem Formulation}
\paragraph{Latent-Space Communication.} Latent-space reasoning has recently gained attention as a promising alternative to text-space reasoning \citep{hao2024training, zhu2025survey}, which is inherently limited by the constraints of natural language. In text space, the reliance on discrete tokens and natural language syntax restricts the expressive bandwidth of models. 
Recent work \citep{zhu2025reasoning} demonstrates that latent representations encode richer information than discrete token outputs. For instance, the Chain of Continuous Thought framework \citep{hao2024training}, which performs multi-step inference within the continuous hidden state space, achieves improved accuracy by enabling more informative intermediate representations. Motivated by these advances, we investigate latent representations as a communication medium between agents. Specifically, we explore continuous latent communication to enhance inter-agent collaboration between DDLMs and ARMs.

Formally, consider two agents in a MAS, where the output of \textit{Agent 1} is fed as input to \textit{Agent 2}. Let the hidden state in step \(t\) in \textit{Agent 1} be denoted by \(\mathbf{h}_t^{(1)} \in \mathbb{R}^d\), where \(d\) is the embedding dimension. In traditional ARMs, recent approaches \citep{hao2024training} use the final hidden state \(\mathbf{h}_T^{(1)}\) as input embedding to predict the \((T+1)\)-th token. In a MAS setting, this would be equivalent to passing \(\mathbf{h}_T^{(1)}\) as an input embedding to \textit{Agent 2}, i.e.,
\[
\mathbf{x}_0^{(2)} = \mathbf{h}_T^{(1)},
\]
where \(\mathbf{x}_0^{(2)}\) is the initial input embedding for \textit{Agent 2}. However, as we explain next, this straightforward approach is not feasible.

\paragraph{Challenge : Embedding Space Mismatch.}
A key challenge arises in directly passing \(\mathbf{h}_T^{(1)}\) from the DDLM (\textit{Agent 1}), which is trained bidirectionally through masked denoising, to the ARM (\textit{Agent 2}), which is trained unidirectionally in an autoregressive manner. These fundamentally different training paradigms result in different embedding spaces. Specifically, the hidden representations \(\mathbf{h}_T^{(1)}\) and \(\mathbf{h}_0^{(2)}\) lie in separate latent manifolds:
\[
\mathbf{h}_T^{(1)} \in \mathcal{H}_\text{DDLM} \subseteq \mathbb{R}^{d_1}, \quad
\mathbf{h}_0^{(2)} \in \mathcal{H}_\text{ARM} \subseteq \mathbb{R}^{d_2},
\]
with typically \(d_1 \neq d_2\) and differing geometric and statistical properties.
Consequently, a direct assignment \(\mathbf{x}_0^{(2)} = \mathbf{h}_T^{(1)}\) is not feasible, motivating the need for a learned or engineered mapping between these heterogeneous latent spaces to enable effective inter-agent communication.

\begin{figure*}[t]
    \centering
    \includegraphics[width=0.75\textwidth]{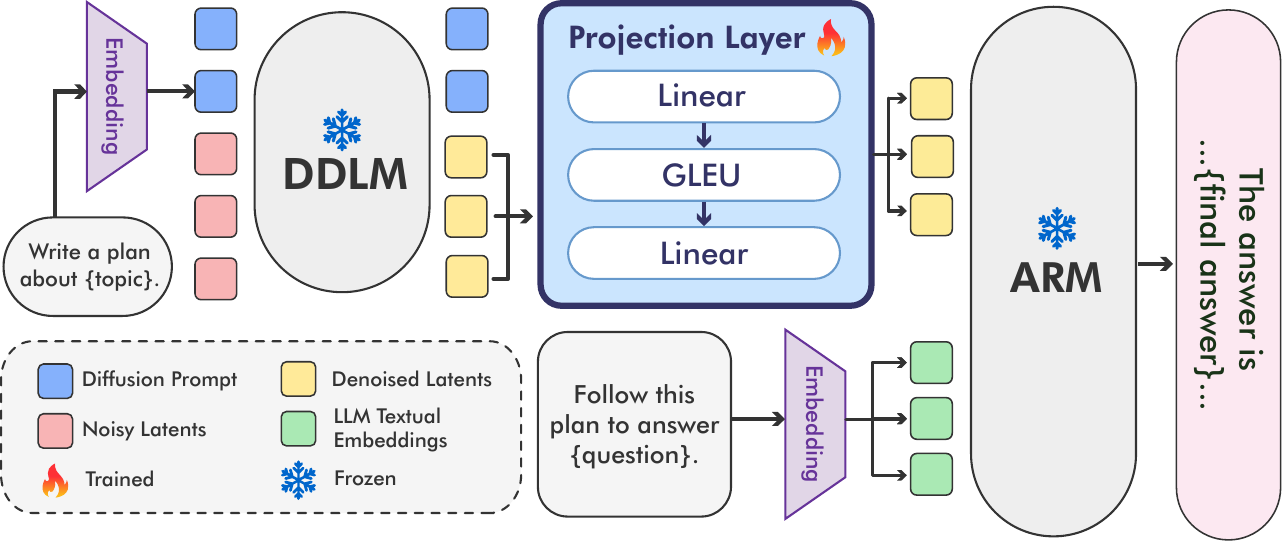}
    \caption{Overview of the latent-space collaboration pipeline. A discrete diffusion language model (DDLM) generates a latent plan. The plan is projected directly into the autoregressive model (ARM) embedding space through a learned projector (in blue). The ARM then conditions on the plan and the question to produce the final answer.
    }
    \label{fig:projector}
\end{figure*}

To address this, we propose learning a dedicated projection network (Figure~\ref{fig:projector}) that maps latent representations from the DDLM space to the ARM space, thereby allowing meaningful translation and collaboration despite architectural and representational heterogeneity.

\section{Latent-DARM}
\label{sec:methods}

\subsection{System Architecture}
We compare the following two planner–executor integration schemes.

\paragraph{Text-Space Interface (Baseline).}
The conventional planner–executor interface operates in text space, where the planner’s latent representation is first decoded into a discrete sequence and then re-encoded by the executor:
\begin{equation}
h_{\text{DDLM}}
\xrightarrow{\pi_{\text{decode}}} T
\xrightarrow{\phi_{\text{encode}}} h_{\text{ARM}}.
\end{equation}

\paragraph{Latent-DARM (Proposed).}
In contrast, we propose a direct latent-space interface that bypasses explicit text generation:
\begin{equation}
h_{\text{DDLM}} \xrightarrow{f_\theta} h_{\text{ARM}},
\end{equation}
where $f_\theta : \mathcal{H}_{\text{DDLM}} \rightarrow \mathcal{H}_{\text{ARM}}$ is a learned projection module implemented as a Linear–GELU–Linear network, as shown in Figure~\ref{fig:projector}. The projection is trained to align the planner and executor representation spaces, enabling information transfer without intermediate discretizations.

\paragraph{Remark.}
Here, the only trainable component is the projector, which learns to map latents from the last hidden layer after the final denoising step of the DDLM (\emph{Agent 1}) to an input embedding. This embedding is then concatenated with the prompt embedding of the \emph{Agent 2} ARM. 
It is important to note that we do \emph{not} perform any fine-tuning of the agents themselves.

\subsection{Projector Training}

We train the projection module $f_\theta$ while keeping both the DDLM planner and the ARM executor frozen.
A natural but problematic approach would be to directly align the projected planner representation with an
``ideal'' executor representation via a distance-based objective,
\begin{equation}
\min_\theta \; \mathbb{E}\big[ \| f_\theta(h_{\text{DDLM}}) - h_{\text{ARM}}^\ast \|_2^2 \big],
\end{equation}
where $h_{\text{ARM}}^\ast$ denotes a target embedding in the executor space.
However, such an objective is ill-defined in practice: the executor does not admit a unique canonical hidden
state corresponding to a correct solution. Furthermore, defining $h_{\text{ARM}}^\ast$ via text-based execution would
reintroduce the very discretization bottleneck that our latent interface is designed to avoid.

As an alternative to this approach, which is one of the key contributions of this work, we adopt a training objective task-based that optimizes the projection indirectly through downstream
performance. 
Concretely, given a dataset of reasoning tasks comprising question-answer pairs, for each input question \( q \) with ground-truth answer \( a \), we extract the latent representation of the planner \( h_{\mathrm{DDLM}}(q) \) from the final denoising step. This latent state is then projected through \( f_\theta \) and concatenated with the embedding of the encoded question to condition the frozen ARM executor. The projector is trained by minimizing the negative log-likelihood of the
correct answer:
{
\setlength{\abovedisplayskip}{5pt}
\setlength{\belowdisplayskip}{5pt}
\begin{equation}
\min_\theta \; \mathbb{E}_{(q,a)\sim\mathcal{D}}
\big[ -\log p_{\text{ARM}}\big(a \mid f_\theta(h_{\text{DDLM}}(q)), q \big) \big].
\end{equation}
}
Rather than imposing
geometric proximity to an ill-defined target embedding, it encourages $f_\theta$ to map planner latents into
regions of the executor’s representation space that induce similar downstream behavior. In other words,
the projection is optimized to preserve functional equivalence with respect to the task, rather than
geometric similarity in the embedding space. Through backpropagation, the gradients in the executor’s output
implicitly shape the projection to achieve optimal task alignment, without requiring intermediate text
generation or oracle supervision.
The training procedure for the projector is described in Algorithm~\ref{alg:training}.

\begin{algorithm}
\caption{Latent-DARM Training}
\label{alg:training}
\textbf{Input:} data $\mathcal{D}$, planner $\theta_{\text{DDLM}}$, executor $\theta_{\text{ARM}}$, projection $f_\theta$, rate $\eta$, epochs $E$, size $B$
\begin{algorithmic}[1]
\STATE Initialize $\theta$ randomly
\STATE \textbf{for} each epoch \textbf{do}
\STATE ~~~\textbf{for} each batch \textbf{do}
\STATE ~~~~~~\textbf{for} $i$ in batch \textbf{do}
\STATE ~~~~~~~~~$h_i^{\text{DDLM}} = \text{DDLM}_{\text{planner}}(q_i)$
\STATE ~~~~~~~~~$h_i^{\text{proj}} = f_\theta(h_i^{\text{DDLM}})$
\STATE ~~~~~~~~~$h_i^{\text{ARM}} = [h_i^{\text{proj}}; \text{embed}_{\text{ARM}}(q_i)]$
\STATE ~~~~~~$\mathcal{L} = -\frac{1}{B} \sum_{i=1}^{B} \log p_{\text{ARM}}(a_i \mid h_i^{\text{ARM}})$
\STATE ~~~~~~$\theta = \theta - \eta \nabla_\theta \mathcal{L}$
\end{algorithmic}
\end{algorithm}

\subsection{Projector Inference and Evaluation}

The projector’s inference procedure is detailed in Algorithm~\ref{alg:inference}.
We conduct experiments on the benchmarks described in Section~\ref{sec:Experimental}.

\begin{algorithm}
\caption{Latent-DARM Inference}
\label{alg:inference}
\textbf{Input:} query $q$, projection $f_{\theta^*}$, planner, executor
\begin{algorithmic}[1]
\STATE $h^{\text{DDLM}} = \text{DDLM}_{\text{planner}}(q)$
\STATE $h^{\text{proj}} = f_{\theta^*}(h^{\text{DDLM}})$
\STATE $h^{\text{input}} = [h^{\text{proj}}; \text{embed}_{\text{ARM}}(q)]$
\STATE $a = \text{ARM}_{\text{executor}}(h^{\text{input}})$
\STATE \textbf{return} $a$
\end{algorithmic}
\end{algorithm}

\section{Experimental Evaluation}
\label{sec:Experimental}
\subsection{Models and Benchmarks}

\paragraph{DDLMs.}  
We use two DDLMs, LLada-8B-Instruct \citep{nie2025largelanguagediffusionmodels} and Dream-v0-Instruct-7B \citep{ye2025dream7bdiffusionlarge}. The default sequence length is set to 128 tokens, which provides sufficient capacity for plan generation while reducing repetition errors (see Appendix~\ref{sec:planner_repetition}).

\textbf{ARMs.}
\textbf{Non-reasoning models:}
We consider two ARMs of comparable scale—Qwen2.5-7B-Instruct \citep{qwen25} and Llama-3.1-8B-Instruct \citep{touvron2023llamaopenefficientfoundation, dubey2024llama}—for fair comparison. We also include smaller variants, Llama-3.2-3B-Instruct and Qwen2.5-3B-Instruct \citep{qwen25}, to assess performance in lower-capacity regimes.
\textbf{Reasoning models:}
We also evaluate Qwen3-1.7B \citep{qwen3} and DeepSeek-R1-Distill-Qwen-7B \citep{qwen25}, a distilled variant that compresses the reasoning capabilities of DeepSeek-R1 into a Qwen-7B backbone. For simplicity, we refer to this model as the \emph{DeepSeek} model. Our goal is not to match specialized reasoning systems, but to contextualize hybrid ARM–DDLM collaboration against strong reasoning baselines.

\textbf{Benchmarks.} We evaluate on a diverse suite of reasoning benchmarks: ARC-E and ARC-C \citep{Clark2018ThinkYH}, science exam questions at Easy and Challenging difficulty; MMLU \citep{hendryckstest2021,hendrycks2021ethics}, which spans mathematics, history, computer science, and law; and AIME 2024 \citep{aime_1983_2024,aime_2024}, a high-school mathematics competition. We also include DART-1 through DART-5 \citep{tong2024dartmathdifficultyawarerejectiontuning}, a large-scale mathematical reasoning benchmark covering five difficulty levels. These benchmarks are standard in the literature and together provide broad coverage of reasoning domains. For each evaluation, we used the largest of 200 samples or the full benchmark size.

\paragraph{Projector Training Setup.}
We employ Latent-DARM equipped with a custom latent projector illustrated in Figure~\ref{fig:projector}, consisting of three linear layers interleaved with two GELU activations. We train three projector variants, each using a different DDLM latent sequence length (64, 128, and 256). The projector maps a 4096-dimensional input to a 1024-dimensional bottleneck and subsequently projects to LLaMA's hidden dimension, followed by LlamaRMSNorm normalization.
The training data comprises latent representations of shape $\{64,128,256\} \times 4096$ tokens, extracted from the last hidden layer of the DDLM LLaDA-8B-Instruct model after the final denoising step. This dataset contains 35,000 samples uniformly drawn from seven datasets: ARC\_Easy, ARC\_Challenge, and DART1--DART5 (5,000 samples each).
The ARM model used for training is Llama-3.2-3B-Instruct using bfloat16 precision. Training utilizes the AdamW optimizer (PyTorch) with a learning rate of $5 \times 10^{-4}$, weight decay of 0.001, and 300 warmup steps, combined with cosine learning rate scheduling. We use a batch size of 4 per device, with gradient accumulation over 2 steps (effective batch size 8), training for 10 epochs.
LoRA adapters (rank 8, alpha 32) are applied to the executor's attention and feed-forward network (FFN) projection layers for computational efficiency, while the LLaMA backbone and language 
modeling head remain frozen. 

\section{Results and Discussion}

\subsection{Results}

\begin{table*}[t]
\centering
\caption{Evaluation of DeepSeek-R1-Distill-Qwen-7B and Qwen3-1.7B on reasoning benchmarks, including text-space vs. latent-space collaboration.}
\label{tab:model_performance}
\renewcommand{\arraystretch}{1.1}
\setlength{\tabcolsep}{5pt}
\resizebox{\textwidth}{!}{%
\begin{tabular}{lccccccccc}
\toprule
\rowcolor{gray!10}\textbf{Model / Setting} & \textbf{ARC-E} & \textbf{ARC-C} & \textbf{DART-1} & \textbf{DART-2} & 
\textbf{DART-3} & \textbf{DART-4} & \textbf{DART-5} & \textbf{AIME24} & \textbf{MMLU} \\
\midrule

\rowcolor{gray!20}\multicolumn{10}{c}{\textbf{Accuracy}} \\

DeepSeek-R1 (ARM only ) & 94.0 & 88.0 & 89.0 & 81.5 & 85.5 & 75.0 & 61.5 & 28.5 & 60.0 \\

Qwen3-1.7B (ARM only) & 92.0 & 86.0 & 89.5 & 86.0 & 83.0 & 73.0 & 49.0 & 8.5 & 68.5 \\

Llama-3.2-3B (ARM only) & 87.5 & 79.5 & 44.5 & 35.5 & 28.0 & 34.5 & 36.5 & 3.5 & 54.0 \\

LLaDA-8B $\to$ Llama-3.2-3B (text-space) & 90.5 & 82.5 & 53.5 & 43.0 & 35.5 & 30.0 & 27.0 & 0.0 & 52.5 \\

\rowcolor{yellow!10} LLaDA-8B $\to$ Llama-3.2-3B (latent-space) &   &   &   &   &   &   &   &  &  \\
\rowcolor{yellow!10}\emph{64 tokens plan}            & 85.0 & 78.5 & 78.5 & 62.5 & 57.0 & 63.0 & 54.0 & 12.5 & 52.0 \\
\rowcolor{yellow!10}\emph{128 tokens plan}          & 87.5 & 76.5 & 70.5 & 43.0 & 43.0 & 49.0 & 36.0 & 14.0 & 44.0 \\
\rowcolor{yellow!10}\emph{256 tokens plan  }         & 85.0 & 81.0 & 70.0 & 62.0 & 50.0 & 62.0 & 52.5 & 14.0 & 15.5 \\

\midrule
\rowcolor{gray!20}\multicolumn{10}{c}{\textbf{Number of Tokens in average}} \\

DeepSeek-R1 (ARM only) & 398  & 504  & 1420 & 1833 & 2076 & 2418 & 3068 & 3832 & 760 \\

Qwen3-1.7B (ARM only) & 282  & 397  & 1024 & 1669 & 2112 & 2550 & 3106 & 4036 & 984 \\

Llama-3.2-3B (ARM only)  & 4 & 4 & 17 & 16 & 22 & 28 & 41 & 462 & 4 \\

LLaDA-8B $\to$ Llama-3.2-3B (text-space) & 4 & 4 & 10 & 14 & 16 & 12 & 20 & 503 & 4 \\

\rowcolor{yellow!10}LLaDA-8B $\to$ Llama-3.2-3B (latent-space) &   &   &   &   &  &   &   &   &  \\

\rowcolor{yellow!10}\emph{plan (64, 128, 256 tokens) + executor :\textsuperscript{†} } & 2 & 2 & 4 & 5 & 5 & 5 & 6 & 14 & 2 \\

\bottomrule \multicolumn{10}{l}{\scriptsize\textsuperscript{†}\textit{Executor token averages; planner tokens (64, 128, 256) are to be added implicitly.}}
\end{tabular}
}
\end{table*}

\begin{figure}[h]
    \centering
    \includegraphics[width=0.35\textwidth]{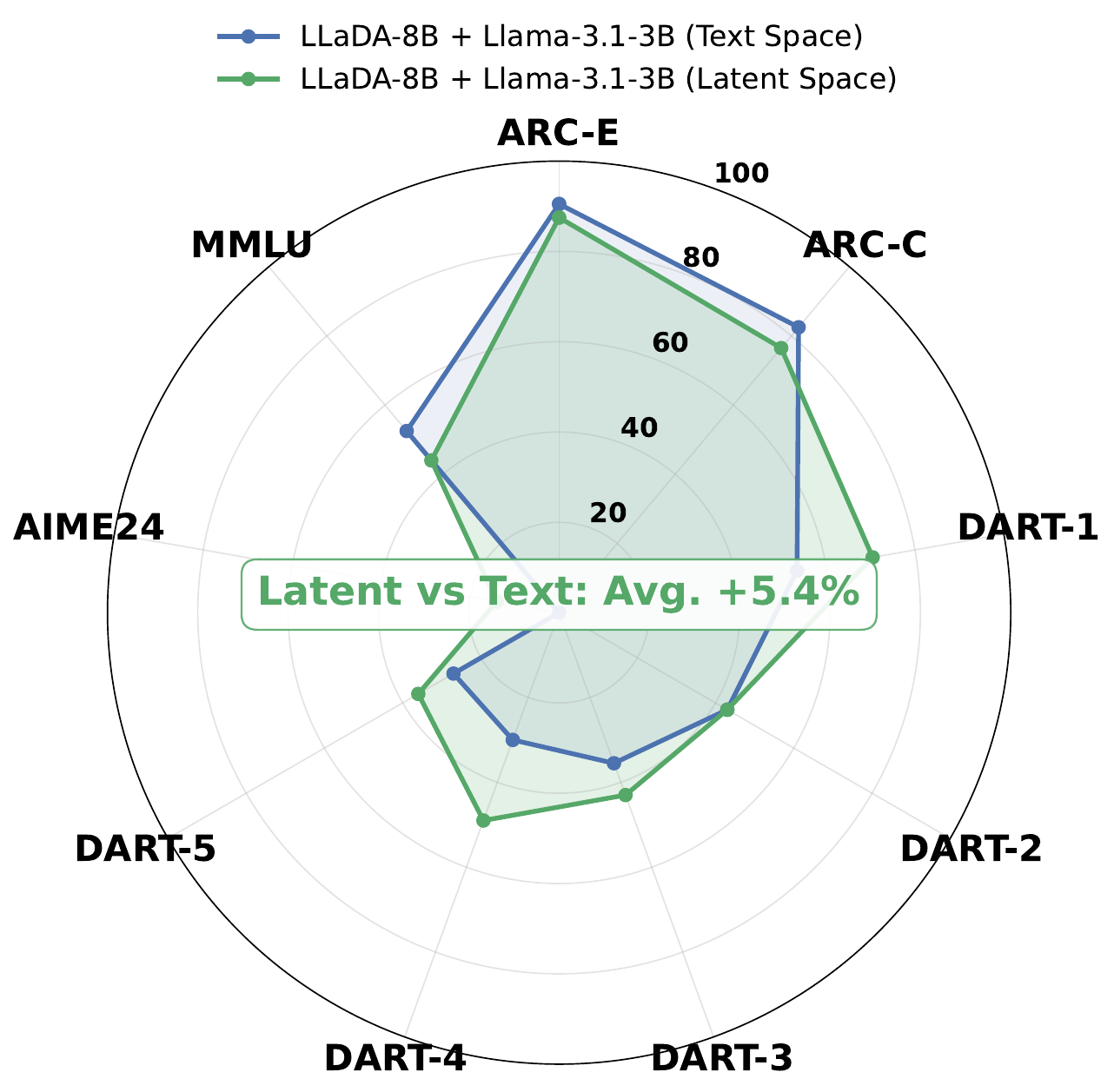}
    \caption{\textbf{Accuracy comparison of text-space vs.\ latent-space collaboration.} }

    \label{fig:latentsbar}
\end{figure}

\paragraph{Accuracy.}
The results of Latent-DARM compared to the text-space baseline are reported in Table~\ref{tab:model_performance} and illustrated in Figure~\ref{fig:latentsbar}. While performance on ARC-E (85.0 vs.~90.5) and ARC-C (81.0 vs.~82.5) remains comparable across the two settings, the latent space consistently yields substantially higher accuracy on the DART benchmarks: e.g., DART-1 (78.5 vs.~53.5), DART-2 (62.5 vs.~43.0), DART-3 (57.0 vs.~35.5), DART-4 (63.0 vs.~30.0), and DART-5 (54.0 vs.~27.0). On AIME, the latent approach performance reaches 12.5\% with the projector trained on 64 tokens and 14\% with the projector trained on 128 or 256 tokens, compared to 0.0\% for the text space. Significantly,this improvement is obtained even though the projector between DDLM and ARM was trained without using any data from AIME or MMLU, yet it still generalizes to deliver markedly stronger results on these challenging evaluations. 
The latent approach underperforms on MMLU (44.0\% vs. 52.5\% text-space), likely because our projector was trained exclusively on reasoning benchmarks (DART, ARC) where planning provides structural guidance. MMLU emphasizes factual recall across diverse domains, where the planning bottleneck may lose fine-grained knowledge while preserving reasoning patterns—a trade-off that benefits multi-step tasks but hinders broad knowledge retrieval.

\textbf{Remark.} \emph{At first glance, a result of 0.0\% may seem surprising. However, when compared to publicly evaluated models such as Mistral-Small-3.1, a 24B parameter model \citep{mistral_small_2503_valsai_2026} achieving 3.54\% on Aime, our result with a 3B model is reasonable.}

\paragraph{Token Budget and Efficiency.}  
We explicitly control the length of diffusion-generated plans (64, 128, 256 tokens). Results demonstrate that longer is not always better:  Notably, the 64-token configuration offers the best overall trade-off between accuracy and efficiency on average. One plausible explanation for this superior performance in MMLU for instance, is its lower degree of redundancy. As shown in Appendix~\ref{sec:planner_repetition}, the repetition rate of diffusion outputs with 64 tokens in LLaDA-8B is comparable to that of Qwen2.5-7B and remains similarly low, whereas configurations with 128 and 256 tokens exhibit slightly higher repetition. 

Most importantly, Latent-DARM is markedly more efficient than both baselines 
(DDLM$\to$ARM in text space and ARM-only) as well as reasoning models. 
Remarkably, with only 64 planner tokens and an average of 5 executor tokens, it surpasses \textsc{Qwen3} on DART-5 while using merely \textbf{2.2\%} of the tokens, and outperforms it on AIME with just \textbf{1.9\%}. 
Although it does not yet reach the raw accuracy of \textsc{DeepSeek-R1}, it achieves highly competitive performance at a fraction of the computational cost, 
as reflected in the average token usage reported in Table~\ref{tab:model_performance}.

\begin{figure*}
    \centering
    \includegraphics[width=\textwidth]{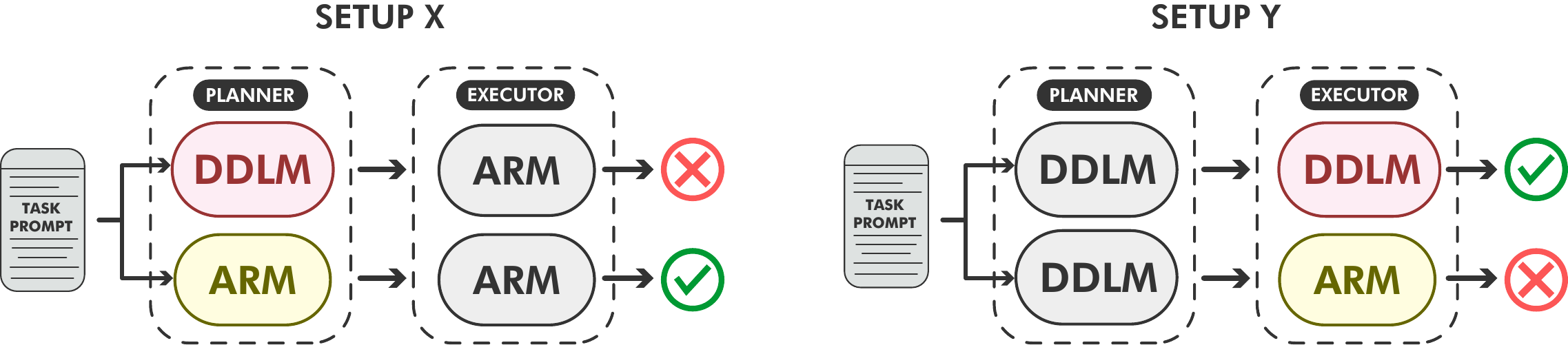}
    \caption{Diagnostic configurations for attributing errors to planner or executor. 
    \textbf{Setup X} tests whether failures stem from the planner: if replacing the diffusion planner (DDLM) with an autoregressive planner (ARM) fixes the output, 
    the error is attributed to the DDLM. 
    \textbf{Setup Y} tests executor reliability: if a diffusion executor succeeds where an ARM executor fails, the limitation lies in the executor.}
    \label{fig:setupxy}
\end{figure*}

\subsection{Are we improving the planner ?}
\label{sec:diagnosing_failures}

To determine whether the gains from latent-space collaboration stem from improved plan communication—our central motivation—we perform a diagnostic analysis that disentangles planner and executor failures in DDLM$\rightarrow$ARM collaboration. For each interface (text-space and latent-space), we consider two subsampled setups. \textbf{Setup X} consists of questions where DDLM$\rightarrow$ARM fails but ARM$\rightarrow$ARM succeeds, indicating a failure of the DDLM planner, since the executor is capable of solving the task when provided with a coherent plan. \textbf{Setup Y} consists of questions where DDLM$\rightarrow$DDLM succeeds but DDLM$\rightarrow$ARM fails, indicating a limitation of the executor, as the planner output is sufficient and the executor is the only component that changes. We define a failure as a question for which the accuracy is zero. This diagnostic allows us to attribute DDLM$\rightarrow$ARM failures to either the planner or the executor and to examine how this attribution shifts when moving from text-space to latent-space collaboration.

\begin{figure}[htbp!]
  \centering
  \includegraphics[width=0.8\linewidth]{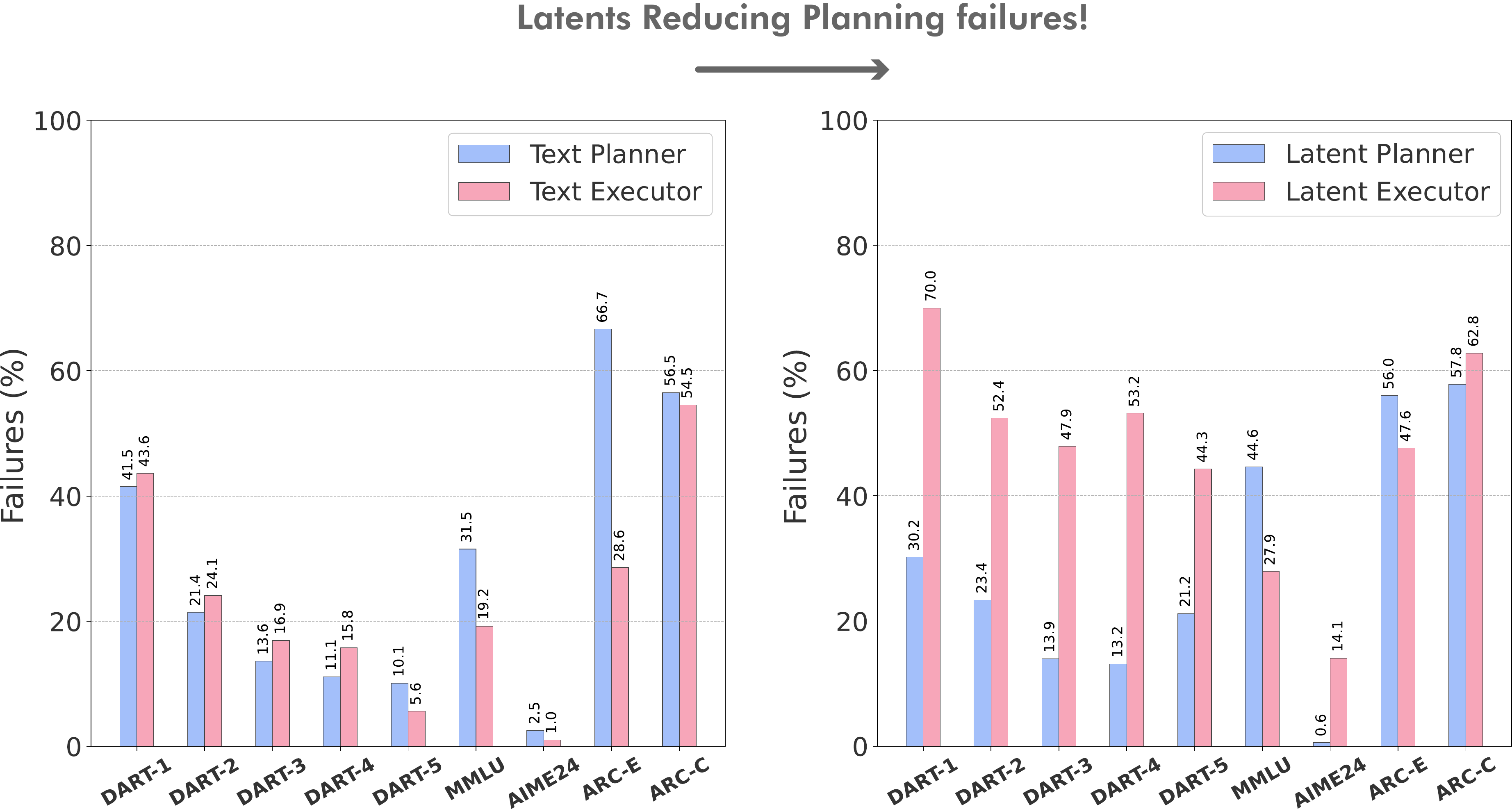}
  \caption{\textbf{Planner vs.\ executor failures in text- vs.\ latent-space collaboration.} 
  Results for LLaDA-8B-Instruct and Llama-3.2-3B-Instruct. 
  Latent-space collaboration substantially reduces planning errors compared to text-space.}
  \label{fig:performance_comparison}
\end{figure}

We quantify these effects using $\textbf{Percentage}_X$, the fraction of DDLM$\rightarrow$ARM failures attributable to the planner (Setup X), and $\textbf{Percentage}_Y$, the fraction attributable to the executor (Setup Y), defined as
\vspace{-0.5pt}
\begin{align*}
\text{\textbf{Percentage}}_{i} &= 
\frac{\# \{\text{Setup } i \text{ samples}\}}
{\# \{\text{Incorrect samples in } \text{DDLM} \to \text{ARM}\}} \\
&\quad \text{for } i \in \{X, Y\}.
\end{align*}
Figure~\ref{fig:setupxy} illustrates these two setups. In both setups, ARM$\rightarrow$ARM and DDLM$\rightarrow$DDLM are always evaluated in text space; only DDLM$\rightarrow$ARM is varied between text and latent communication, isolating the effect of the interface. As shown in Figure~\ref{fig:performance_comparison}, under text-space collaboration, most failures fall into Setup X, indicating that performance is primarily limited by planning degradation induced by textual decoding. In contrast, under Latent-DARM, failures shift predominantly to Setup Y, demonstrating that latent-space communication substantially improves planning fidelity, with the executor emerging as the dominant bottleneck. Exact failure distributions are reported in Appendix~\ref{appendix:comparison}.

\textbf{Remark.} This observation does \emph{not} imply that the executor makes more mistakes. The reported quantities are relative percentages over failures: the shift indicates that, under Latent-DARM, a smaller fraction of errors is attributable to the planner. Consequently, the observed accuracy gains can be attributed to improved planner communication.

\section{Conclusion}

We introduced \textbf{Latent-DARM}, a latent-space collaboration framework that enables effective cooperation between discrete diffusion language models (DDLMs) and autoregressive language models (ARMs) in multi-agent reasoning systems. By replacing text-based interfaces with a learned latent projection, our approach allows diffusion planners to transmit globally structured planning information to autoregressive executors without being constrained by fluency limitations.

Empirically, latent-space communication outperforms text-based collaboration on average, particularly on planning-intensive benchmarks such as DART and AIME. Diagnostic analyses show that these gains primarily arise from a reduction in planning failures, indicating that latent exchange preserves high-level reasoning structure that is otherwise degraded by textual decoding. Moreover, Latent-DARM achieves competitive performance with state-of-the-art reasoning models while using orders of magnitude fewer tokens, demonstrating that strong reasoning does not require long textual chains of thought.

Overall, this work suggests that text does not need to be the sole medium of inter-agent communication. Latent-space interfaces provide a high-bandwidth, task-aligned alternative that enables efficient collaboration between heterogeneous models, opening new directions for scalable and budget-aware reasoning systems.

\paragraph{Future Work.} Our results suggest that latent-space communication represents an underexplored 
but promising paradigm for multi-agent systems. The dramatic efficiency gains and improvements on planning-intensive tasks indicate this approach deserves further systematic investigation. Key directions include: developing 
adaptive architectures that route between latent and text modes based on task characteristics, 
scaling to diverse domains to understand generalization limits, extending to bidirectional 
and multi-hop agent collaboration, and establishing theoretical foundations for when and why 
latent communication succeeds. More broadly, this work challenges the assumption that natural 
language is the optimal inter-agent medium, opening questions about what other structured 
representations might enable even tighter model integration.

\bibliography{iclr2026_conference}
\bibliographystyle{iclr2026_conference}

\appendix
\section*{Appendix}

\section{More Insights On Planner Repetition of tokens}
\label{sec:planner_repetition}

We will perform a qualitative assessment to identify prompt repetition errors in the planner text in the setup DDLM $\to$ ARM in the text space, using 256, 128 and 64 tokens for the LlaDa-8B-Instruct model, as well as Qwen2.5-7B-Instruct and Dream-v0-7B-Instruct for comparison. This analysis examines how increasing the number of planning tokens affects repetition, relative to an autoregressive model baseline.

We use the following metrics proposed in \citep{dragon}:
\begin{itemize}
    \item Distinct-3 (D-3)
    \item Repetition-4 (R-4)
    \item Lexical Repetition (LR-n)
    
\end{itemize}
\vspace{0.5cm}

\textbf{Distinct-3 (D-3)} calculates the percentage of unique 3-grams over all 3-grams. The value of Distinct-3 takes values between 0 and 1, with the closer to 1 indicating that the text is more diverse at the 3-gram level. Let \( D_3 \) be the number of unique 3-grams in the text and \( T_3 \) be the total number of 3-grams in the text. Distinct-3 is then computed by the following formula:

\[
\text{Distinct-3} = \frac{D_3}{T_3} \times 100
\]

\textbf{Repetition-4(R-4)} 
Let \( T \) be the total number of sentences in the text, \( R_t \) be the number of 4-grams repeated in a sentence \( t \), and \( I(x) \) be an indicator function (1 if \( x \) is true, 0 if \( x \) is false). Then Repetition-4 is calculated as follows:

\[
\text{Repetition-4} = \frac{1}{T} \sum_{t=1}^{T} I(R_t > 1) \times 100
\]

\textbf{Lexical Repetition (LR-n)} computes the average percentage of 4-grams that occur at least \( n \) times in the generated text. Let \( G \) be the total number of possible 4-grams in all texts and \( L_g \) be the number of repetitions of \( G \), then Lexical Repetition (LR-n) is calculated by the following formula:

\[
\text{Lexical Repetition} = \frac{1}{G} \sum_{g=1}^{G} I(L_g \geq n) \times 100
\]

\paragraph{Results}

\begin{table*}[ht]
\centering
\caption{Repetition Evaluation}
\small
\begin{tabular}{|l|c|c|c|}
\hline
\textbf{Simulation} & \textbf{D-3} & \textbf{R-4} & \textbf{LR-n} \\
\hline
  & $\uparrow$ & $\downarrow$ & $\downarrow$ \\
\hline
 \texttt{Qwen2.5-7B-Instruct (Baseline)} & 98.47   & 3.05  & 0.64 \\
\texttt{LLaDA-8B-Instruct (256 tokens)} & 83.05 & 10.52 & 6.74 \\
\texttt{LLaDA-8B-Instruct (128 tokens)} & 93.15  &  5.33 &  3.43 \\
\texttt{LLaDA-8B-Instruct (64 tokens)} &  96.85  & 3.37   & 1.33   \\
\texttt{Dream-v0-7B-Instruct (256 tokens)} &  62.02 & 3.26  & 2.15\\
\hline
\end{tabular}
\label{tab:fluency}
\end{table*}

In table \ref{tab:fluency}, the $\uparrow$ indicates that a larger value corresponds to better performance, while the $\downarrow$ indicates that a smaller value corresponds to better performance.

We observe a tendency for greater repetition in the plans generated by LLADA as the number of tokens increases.The number of repetitions in diffusion with 64 tokens in LLADA is comparable to that of Qwen2.5-7B and remains similarly low.

\section{Prompts for LLM Experiments}
\label{appendix:prompts}

\subsection{Planner Prompt}
{
\begin{verbatim}
You are a careful problem-solving planner.

Task: Produce ONLY a short list of HINTS that help
solve the question. 
Do NOT state or imply the final answer. 
Do NOT mention any option letter (A, B, C, or D). 
Do NOT quote any option text verbatim. 
If you find yourself about to reveal a specific
option or an answer, replace it with “[HIDDEN]”.

Format:
- Key facts to recall (2–4 bullets)
- Reasoning steps or elimination rules (2–5 bullets)
- Useful equations or definitions (if relevant)
- Edge cases or common traps (optional)

Be concise (<=120 words). No “Answer:” line. 
No letters A–D.

Question (stem only):
{question}
\end{verbatim}
}

\subsection{Executor Prompt}
{
\begin{verbatim}
You are an expert in solving multiple-choice questions.
Given the following plan or reasoning, please solve the question. 
If the plan contains any explicit answer or option letter, 
ignore it and solve from the hints + question only.

Plan:
{plan}
{question}
\end{verbatim}
}

\section{Performance Comparison of Planner vs.\ Executor Issues}
\label{appendix:comparison}
\begin{table}[H]
\centering
\caption{Performance comparison of planner vs.\ executor issues for LLaDA-8B-Instruct and Llama-3.2-3B-Instruct under \textbf{Text-Space} vs.\ \textbf{Latent-Space} collaboration.}
\small
\renewcommand{\arraystretch}{0.5}
\begin{tabular}{l>{\columncolor{cyan!20}}c>{\columncolor{orange!20}}c>{\columncolor{gray!15}}c}
\toprule
\rowcolor{navy!80}
\textcolor{white}{\textbf{Benchmark}} & 
\textcolor{white}{\textbf{\begin{tabular}{@{}c@{}}Planning\\Failures (\%)\end{tabular}}} & 
\textcolor{white}{\textbf{\begin{tabular}{@{}c@{}}Execution\\Failures (\%)\end{tabular}}} & 
\textcolor{white}{\textbf{\begin{tabular}{@{}c@{}}Error Gap\\(\%)\end{tabular}}} \\
\midrule

\multicolumn{4}{l}{\textbf{ \textcolor{blue!80!black}{LLaDA-8B + Llama-3.2-3B (Text-Space Pipeline)}}} \\
\midrule
DART-1        & 41.50 & \cellcolor{red!10}\textbf{43.64} & 2.14 \\
DART-2        & 21.43 & \cellcolor{red!10}\textbf{24.14} & 2.71 \\
DART-3        & 13.60 & \cellcolor{red!10}\textbf{16.92} & 3.32 \\
DART-4        & 11.11 & \cellcolor{red!10}\textbf{15.79} & 4.68 \\
DART-5        & \cellcolor{blue!10}\textbf{10.12} & 5.63 & 4.49 \\
MMLU          & \cellcolor{blue!10}\textbf{31.52} & 19.23 & 12.29 \\
AIME24        & \cellcolor{blue!10}\textbf{2.54}  & 1.03 & 1.51 \\
ARC-E         & \cellcolor{blue!10}\textbf{66.67} & 28.57 & 38.10 \\
ARC-C         & \cellcolor{blue!10}\textbf{56.52} & 54.55 & 1.97 \\
\midrule

\multicolumn{4}{l}{\textbf{ \textcolor{purple!80!black}{LLaDA-8B + Llama-3.2-3B (Latent-Space Pipeline)}}} \\
\midrule
DART-1        & 30.23 & \cellcolor{red!10}\textbf{70.00} & 39.77 \\
DART-2        & 23.37 & \cellcolor{red!10}\textbf{52.42} & 29.05 \\
DART-3        & 13.95 & \cellcolor{red!10}\textbf{47.88} & 33.93 \\
DART-4        & 13.15 & \cellcolor{red!10}\textbf{53.19} & 40.04 \\
DART-5        & 21.21 & \cellcolor{red!10}\textbf{44.29} & 23.08 \\
MMLU          & \cellcolor{blue!10}\textbf{44.64} & 27.90 & 16.74 \\
AIME24        & 0.58  & \cellcolor{red!10}\textbf{14.07} & 13.49 \\
ARC-E         & \cellcolor{blue!10}\textbf{56.00} & 47.61 & 8.39 \\
ARC-C         & 57.80 & \cellcolor{red!10}\textbf{62.79} & 4.99 \\
\bottomrule
\end{tabular}
\end{table}

\section{Detailed results from experiments}

\begin{table}[H]
\centering
\caption{Consolidated evaluation results on text-space collaboration across all model combinations and reasoning benchmarks. $\dagger$ Evaluated with \texttt{enable\_thinking=True},  $\ddagger$ with \texttt{enable\_thinking=False}.}
\label{tab:consolidated_performance}
\renewcommand{\arraystretch}{1.2}
\setlength{\tabcolsep}{1pt}
\rowcolors{3}{gray!5}{white}
\resizebox{\textwidth}{!}{%
\begin{tabular}{llccccccccc}
\toprule
\rowcolor{gray!10} \textbf{Model / Combination} & \textbf{Setup} & \textbf{ARC-E} & \textbf{ARC-C} & 
\textbf{DART-1} & \textbf{DART-2} & \textbf{DART-3} & \textbf{DART-4} & 
\textbf{DART-5} & \textbf{AIME} & \textbf{MMLU} \\

\midrule
\rowcolor{gray!20}\multicolumn{11}{c}{\textbf{ARMs only}} \\
\midrule

{Qwen2.5-3B} & ARM  & 91.0 & 86.5 & 58.0 & 34.0 & 29.5 & 17.5 & 11.5 & 1.0 & 61.0 \\
 & ARM $\to$ ARM  & 89.5 & 81.5 & 64.0 & 49.0 & 30.5 & 24.0 & 16.0 & 0.0 & 59.0 \\

{Dream-v0-7B + Qwen2.5-3B} & DDLM $\to$ ARM & 92.0 & 86.0 & 65.5 & 40.5 & 31.0 & 25.0 & 12.5 & 0.5 & 60.5 \\

{LLaDA-8B + Qwen2.5-3B} & DDLM $\to$ ARM & 88.5 & 85.0 & 68.0 & 44.5 & 29.0 & 23.0 & 14.0 & 0.5 & 55.5 \\
 
\midrule
\midrule

{Qwen2.5-7B} & ARM & 94.5 & 90.5 & 68.0 & 48.5 & 35.0 & 35.0 & 19.5 & 2.5 & 67.0 \\
 & ARM $\to$ ARM  & 96.5 & 89.0 & 72.5 & 56.5 & 35.0 & 33.5 & 20.0 & 3.0 & 61.0 \\

{Dream-v0-7B + Qwen2.5-7B} & DDLM $\to$ ARM & 95.5 & 88.5 & 60.5 & 42.0 & 34.0 & 28.5 & 16.0 & 1.0 & 64.5 \\

{LLaDA-8B + Qwen2.5-7B} & DDLM $\to$ ARM & 92.5 & 88.5 & 73.5 & 58.0 & 37.5 & 37.0 & 16.0 & 1.5 & 54.0 \\
 
\midrule
\midrule

{Llama-3.2-3B} 
 & ARM  & 87.5 & 79.5 & 44.5 & 35.5 & 28.0 & 34.5 & 36.5 & 3.5 & 54.0 \\
 & ARM $\to$ ARM  & 91.0 & 80.0 & 47.5 & 37.0 & 31.0 & 30.5 & 30.0 & 0.0 & 56.5 \\

 {Dream-v0-7B + Llama-3.2-3B} & DDLM $\to$ ARM & 91.0 & 79.0 & 54.0 & 39.5 & 32.0 & 31.5 & 27.5 & 0.0 & 59.5 \\

 {LLaDA-8B + Llama-3.2-3B} & DDLM $\to$ ARM & 90.5 & 82.5 & 53.5 & 43.0 & 35.5 & 30.0 & 27.0 & 0.0 & 52.5 \\

 \rowcolor{yellow!10} LLaDA-8B $\to$ Llama-3.2-3B (latent-space) && 87.5 & 76.5 & 70.5 & 43.0 & 43.0 & 49.0 & 36.0 & 14.0 & 44.0 \\
\midrule
\midrule

{Llama-3.1-8B} & ARM  & 87.5 & 79.5 & 68.5 & 50.5 & 36.0 & 36.0 & 20.5 & 2.5 & 63.5 \\
 & ARM $\to$ ARM  & 91.0 & 82.0 & 74.0 & 58.5 & 35.5 & 35.0 & 20.5 & 3.0 & 64.0 \\

 {Dream-v0-7B + Llama-3.1-8B} & DDLM $\to$ ARM & 94.0 & 86.5 & 55.5 & 33.0 & 37.0 & 28.0 & 26.0 & 0.5 & 64.0 \\

 {LLaDA-8B + Llama-3.1-8B} & DDLM $\to$ ARM & 91.5 & 82.5 & 74.0 & 60.0 & 38.5 & 37.0 & 16.5 & 1.5 & 56.5 \\

\midrule
\midrule
{DeepSeek-R1} & ARM  & 94.0 & 88.0 & 89.0 & 81.5 & 85.5 & 75.0 & 61.5 & 28.5 & 60.0 \\

{LLaDA-8B + DeepSeek-R1} & DDLM $\to$ ARM & 92.5 & 88.5 & 73.5 & 58.0 & 37.5 & 37.0 & 16.0 & 1.5 & 54.0 \\

\midrule
\midrule

{Qwen3.1-7B}  & ARM$^{\dagger}$& 92.0 & 86.0 & 89.5 & 86.0 & 83.0 & 73.0 & 49.0 & 8.5 & 68.5 \\

{LLaDA-8B + Qwen3.1-7B} & DDLM $\to$ ARM$^{\ddagger}$ & 91.0 & 82.5 & 87.0 & 70.0 & 58.5 & 46.0 & 27.0 & 1.0 & 52.5 \\

\bottomrule

\end{tabular}
}
\end{table}

\end{document}